\documentclass{clv3}

\NewCommandCopy{\cnumdef}{\numdef}
\NewCommandCopy{\endcnumdef}{\endnumdef}
\let\numdef\relax \let\endnumdef\relax

\usepackage{threeparttable,multirow}
\usepackage{booktabs}
\usepackage{subfigure,epsfig,color,longtable}
\usepackage[table,xcdraw]{xcolor}

\usepackage{hyperref}

\definecolor{darkblue}{rgb}{0, 0, 0.5}
\hypersetup{colorlinks=true,citecolor=darkblue, linkcolor=darkblue, urlcolor=darkblue}

\usepackage{xspace}
\usepackage{adjustbox}

\makeatletter
\DeclareRobustCommand\onedot{\futurelet\@let@token\@onedot}%
\def\@onedot{\ifx\@let@token.\else.\null\fi\xspace}%
\def\eg{\emph{e.g}\onedot} %
\def\ie{\emph{i.e}\onedot} %

\makeatother

\usepackage{amsmath} 
\usepackage{empheq}

\newtheorem{thm}{Theorem}[section]

\newtheorem{lem}[thm]{Lemma}

\theoremstyle{plain}

\theoremstyle{plain}
\newtheorem{rem}[thm]{Remark}
\numberwithin{equation}{section}\graphicspath{{~/figures/}}

\newcommand{\vc}[1]{\mathbf{#1}}
\newcommand{\mat}[1]{\mathbf{#1}}
\newcommand{\Real}{\mathbb R}

\newcommand{\be}{\begin{equation}}
\newcommand{\ee}{\end{equation}}

\usepackage{tikz}
\usetikzlibrary{fadings}
\usetikzlibrary{patterns}
\usetikzlibrary{shadows.blur}
\usetikzlibrary{shapes}

\begin{document}

\runningtitle{\texttt{[MASK]} Insertion}
\runningauthor{Yang \emph{et al}}


\newcommand{\ourmethod}{\textit{DDM}\xspace}
\title{Defensive Dual Masking for Robust Adversarial Defense}

\author{Wangli Yang}
\affil{School of Computing and Information Technology, \\
University of Wollongong\\ 
\href{mailto:wangli@uow.edu.au}{wangli@uow.edu.au}
}

\author{Jie Yang\thanks{Corresponding author.}}
\affil{School of Computing and Information Technology, \\
University of Wollongong\\ 
\href{mailto:jiey@uow.edu.au}{jiey@uow.edu.au}
}

\author{Yi Guo}
\affil{School of Computer, Data and Mathematical Sciences, \\
Western Sydney University\\ 
\href{mailto:y.guo@westernsydney.edu.au}{y.guo@westernsydney.edu.au}
}

\author{Johan Barthelemy}
\affil{NVIDIA\\ 
\href{jbarthelemy@nvidia.com}{jbarthelemy@nvidia.com}
}

\maketitle

\begin{abstract}
{
The field of textual adversarial defenses has gained considerable attention in recent years due to the increasing vulnerability of natural language processing (NLP) models to adversarial attacks, which exploit subtle perturbations in input text to deceive models. This paper introduces the Defensive Dual Masking (\ourmethod) algorithm, a novel approach designed to enhance model robustness against such attacks. \ourmethod utilizes a unique adversarial training strategy where \texttt{[MASK]} tokens are strategically inserted into training samples to prepare the model to handle adversarial perturbations more effectively. During inference, potentially adversarial tokens are dynamically replaced with \texttt{[MASK]} tokens to neutralize potential threats while preserving the core semantics of the input. 
The theoretical foundation of our approach is explored, demonstrating how the selective masking mechanism strengthens the model’s ability to identify and mitigate adversarial manipulations. Our empirical evaluation across a diverse set of benchmark datasets and attack mechanisms consistently shows that \ourmethod outperforms state-of-the-art defense techniques, improving model accuracy and robustness. Moreover, when applied to Large Language Models (LLMs), \ourmethod also enhances their resilience to adversarial attacks, providing a scalable defense mechanism for large-scale NLP applications.
} 
\end{abstract}

\section{Introduction}
Language Models (LMs) have significantly advanced the performance of many Natural Language Processing (NLP) tasks, spanning text/document classification, semantic analysis, and topic clustering. However, extensive research has revealed that LMs are susceptible to adversarial attacks, where even subtle perturbations to input texts can adversely affect model performanc. That is, fine-tuned LMs may demonstrate a significant decrease in performance, up to 85\%, due to the presence of even a single-character misspelling within input textse~\cite{DeepWordBug, TextBugger, li2020bert, jin2020bert}, highlighting their limited robustness in generalization. Consequently, there has been a considerable focus on developing adversarial defenses to ensure robust model performance on both original (clean) and adversarial (polluted) inputs.

Research on adversarial defenses have spanned multiple approaches, including data augmentation, model adaptation, and randomized smoothing. 
Data augmentation techniques, commonly referred to as \textit{adversarial training}, introduce controlled perturbations to clean data, generating noisy variants that are used alongside the clean data for model fine-tuning~\cite{yoo-qi-2021-towards-improving, dong2021towards, zhou-etal-2021-defense, li-etal-2021-searching, meng-etal-2022-self}. Although effective, these methods often demand substantial computational resources due to the need for both generating and training on augmented samples. 
Model adaptation approaches concentrate on refining the vanilla model by modifying the training loss function or adjusting the network architecture~\cite{wang2021infobert, le-etal-2022-shield, liu2022flooding}. However, such modifications typically require extensive hyperparameter tuning and are prone to overfitting, thereby potentially undermining the model generalization capabilities.
Another line of work involves ensemble-based randomized smoothing techniques~\cite{ye-etal-2020-safer, zeng-etal-2023-certified}, but these methods incur overhead due to ensemble classification and tend to exhibit inconsistent performance against various attack types~\cite{zhang-etal-2022-improving, xu-etal-2022-towards}. Further elaboration on existing adversarial attack and defense methods is provided in Section~\ref{sec:related}. 
Thus, further investigations are necessary to improve the generalizability and robustness of models against adversarial attacks. 


To take one step towards this goal, this paper introduces a novel adversarial defense algorithm, termed \textbf{D}efensive \textbf{D}ual \textbf{M}asking (\textbf{\ourmethod}). The core of \ourmethod lies in strategically incorporating \texttt{[MASK]} tokens at both the training and inference stages to enhance model robustness. Specifically, during the training stage, random masking is applied to input sequences by directly adding \texttt{[MASK]} tokens rather than replacing existing ones. This simple yet effective technique generates masked variants of the input data, which are then used to fine-tune the vanilla model (eliminating the need for training on the original, unmodified data). During the inference stage, \ourmethod identifies potential adversarial tokens from unseen testing samples and selectively masks them. Notably, our approach does not attempt to predict the content of masked tokens; instead, the masked samples are fed directly into the fine-tuned model for inference. The proposed \ourmethod is characterized by its simplicity and effectiveness across several aspects: 
\begin{enumerate}
    \item Compared to existing data augmentation techniques, which require generating and training on additional samples, \ourmethod merely inserts \texttt{[MASK]} tokens into the original input. Despite the increase in input length, this approach remains computationally cost-effective as it avoids the overhead of generating/training additional data.     
    \item In contrast to model adaptation techniques, \ourmethod preserves the standard model architecture and loss functions. This ensures consistency and compatibility while also simplifying the integration of \ourmethod into existing frameworks, making it highly adaptable in a plug-and-play fashion.     
    \item Finally, unlike randomized smoothing methods that rely on ensemble-based learning, \ourmethod eliminates the need for complex ensembling. This reduction in complexity not only simplifies implementation but also improves the efficiency of \ourmethod across a wide range of applications.
\end{enumerate}

We further observe that a few existing works also employ masking strategies for adversarial defense. However, our proposed method differs from these approaches in several key aspects, as summarized in Table~\ref{tab:maskingComparison}: (1) The \texttt{[Mask]} token is traditionally employed to hold out portions of input tokens for predicting missing tokens, with several studies extending this approach to generate augmented data~\cite{wang-etal-2023-rmlm,li-etal-2023-text,zhao-mao-2023-generative,raman-etal-2023-model,rafiei-asl-etal-2024-robustsentembed}. In contrast, \ourmethod directly utilizes masked token samples during model fine-tuning; and (2) Existing methods often require multiple masked variants of the same input for inference. In \ourmethod, however, no such multiple samples are needed. We employ a simple yet effective masking strategy to eliminate potentially adversarial tokens during inference. 
Empirically, our proposed method, \ourmethod, consistently outperforms recent state-of-the-art baselines across a combination of standard benchmarks and adversarial attack methods. On average, \ourmethod achieves an improvement of average 5.0 absolute points in accuracy.

\begin{table*}[!th]
    \caption{A summary of various masking strategies employed in adversarial defense, where (M) denotes the process of masking a single input several times to generate multiple masked variants.}\label{tab:maskingComparison}
    \centering
    \begin{adjustbox}{max width=0.98\linewidth}
    \begin{tabular}{cccc|ccc}%
    \toprule
        \multirow{2}*{Method} & \multicolumn{3}{c|}{Training} & \multicolumn{3}{c}{Inference} \\ 
        \cmidrule(lr){2-7} & Replace-then-predict & Replace only & Insert  & Replace-then-predict & Replace only & Insert \\ \midrule
	\textbf{RanMASK}~\cite{zeng-etal-2023-certified} & ~ & \checkmark(M) & ~ & ~ & \checkmark(M) & ~   \\ 
	\textbf{RSMI}~\cite{moon-etal-2023-randomized} & ~ & ~ & ~ & ~ & \checkmark(M) & ~   \\ 
	\textbf{RMLM}~\cite{wang-etal-2023-rmlm} & \checkmark & ~ & ~ & \checkmark & ~ & ~   \\ 	
	\textbf{Adv-Purification}~\cite{li-etal-2023-text}& \checkmark & ~ & ~ & ~ & ~ & ~    \\
	\textbf{GenerAT}~\cite{zhao-mao-2023-generative}& \checkmark & ~ & ~ & ~ & ~ & ~    \\	
	\textbf{MVP}~\cite{raman-etal-2023-model}& \checkmark & ~ & ~ & \checkmark & ~ & ~    \\	
	\textbf{MI4D}~\cite{hu-etal-2023-mask} & ~ & ~ & \checkmark & ~ & ~ & \checkmark   \\
	\textbf{{R}obust{S}ent{E}mbed}~\cite{rafiei-asl-etal-2024-robustsentembed} & \checkmark & ~ & ~ & \checkmark & ~ & ~   \\
	\midrule
	\textbf{\ourmethod} & ~ & ~ & \checkmark & ~ & \checkmark & ~  \\
	\bottomrule
    \end{tabular}
    \end{adjustbox}
\end{table*}

The paper is structured as follows: Section~\ref{sec:related} surveys existing work on adversarial attacks and defenses methods. Section~\ref{sec:proposedmethod} introduces the proposed method and offers theoretical analysis into the effects of masked varitions on model fine-tuning and inference. Section~\ref{sec:experiments} evaluates the method across a combination of four highly-competetive benchmarks and four attacking mechanisms, followed by a comprehensive ablation study. Finally, Section~\ref{sec:conclusion} concludes and outlines future research directions.

\section{Related work}\label{sec:related}
As Transformer-based Language Models (LMs) gain widespread use in tasks such as text classification, clustering, and information retrieval, their susceptibility to adversarial attack has become a growing area of research. In this section, we provide a comprehensive review of the existing literature on textual adversarial attacks and defense mechanisms, including techniques for generating adversarial examples at the character and word levels, as well as various strategies proposed to mitigate these attacks and improving model robustness.

\subsection{Adversarial attacks}
Textual adversarial attacks aim to subtly modify input text in ways that cause the target models to produce incorrect predictions, while maintaining coherence with the original content. The effective adversarial examples must satisfy several key criteria~\cite{jin2020bert}: 
\begin{enumerate} 
	\item \textit{Human prediction consistency}: The adversarial examples should not alter human interpretations, resulting in predictions that align with those made for the original inputs. 
	\item \textit{Semantic similarity}: The adversarial examples must preserve the original semantic meaning as perceived by human interpretation. 
	\item \textit{Linguistic fluency}: The adversarial examples should maintain proper grammar and fluency, ensuring that the text remains natural and grammatically correct. 
\end{enumerate}
These conditions ensure that adversarial examples not only deceive the target models but also remain indistinguishable from natural inputs to human observers. Adversarial attacks in the text domain are primarily categorized based on perturbation granularity into two types: \textit{character}-level and \textit{word}-level perturbations.

\noindent \paragraph{\textbf{Character-level attacks}} These attacks primarily manipulate individual characters within words in the original sample. While such perturbations may alter the semantics of the sentence, human prediction remains relatively unaffected to a certain extent due to visual similarity. \textbf{HotFlip}~\cite{ebrahimi-etal-2018-hotflip} represents an early character-based attack method, which utilizes gradients based on a one-hot input representation to identify the potential change with the highest estimated loss. Additionally, this method employs a beam search to find a set of character manipulations to confuse the model. 
\textbf{DeepWordBug}~\cite{DeepWordBug} utilizes four scoring functions to identify crucial words. It subsequently introduces four Token Transformers targeting these significant words, which involve swapping two adjacent letters, substituting a letter with a random one, deleting a random letter, and inserting a random letter. 
Similarly, \textbf{Textbugger}~\cite{TextBugger} initially identifies important words by either computing the Jacobian matrix of the model output or comparing the model changes before and after word deletion. For identified words, Textbugger extends character-level attacks beyond inserting, deleting, and swapping letters by suggesting the replacement of characters with visually similar or adjacent ones from the keyboard.

\noindent \paragraph{\textbf{Word-level attacks}} These attacks deceive models through subtle word manipulations, such as synonym substitution, while maintaining grammatical correctness and semantic similarity. 
\textbf{PWWS}~\cite{ren-etal-2019-generating} employs probability-weighted word saliency to evaluate the sensitivity of the victim model to each input word. Subsequently, candidate words are replaced by their synonyms (from WordNet), taking into account the magnitude of change in the model's output probability. 
\textbf{FGPM}~\cite{wang2020adversarial} begins by constructing a synonym set for each input word using its nearest neighbors in the GloVe vector space. Target words are then selected by evaluating the projected distance and gradient change between the original word and its synonym candidates in the gradient direction. Finally, this method achieves textual attack through word replacement using its synonym set. 
In \textbf{TextFooler}~\cite{jin2020bert}, target words are identified by comparing changes in prediction results before and after a word deletion. Subsequently, TextFooler replaces important words with synonyms that are both semantically similar (minimizing their cosine distance) and grammatically correct (verified through POS checking).  
\textbf{BERT-Attack}~\cite{li2020bert} adopts the Masked Language Modeling (MLM) approach by applying the \texttt{[MASK]} token to replace existing words in the input sentence. Subsequently, the change in output from the victim model serves as an importance score to select target words, which are then replaced by filling the corresponding \texttt{[MASK]} token(s) as part of the MLM process.


\subsection{Adversarial defenses}\label{sec:existingDefenses}
\begin{figure*}[!tph]
	\centering
	\subfigure[data augmentation]{\includegraphics[width=0.315\textwidth] {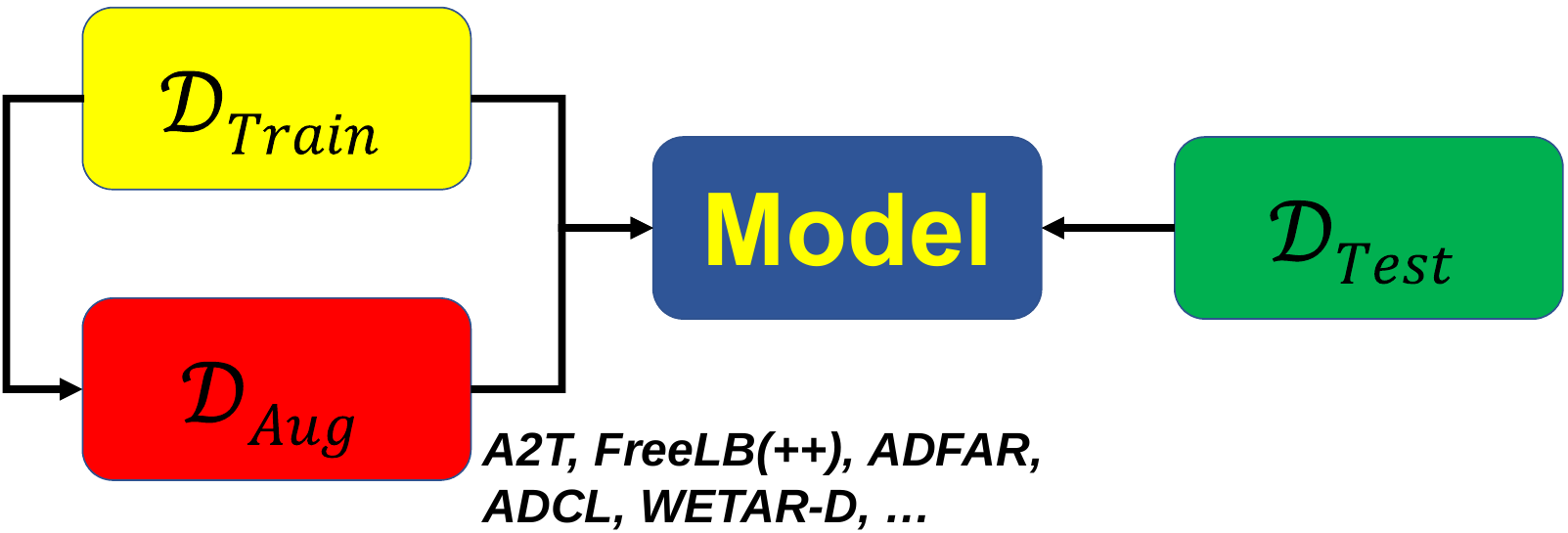}}
~~~\subfigure[model enhancement]{\includegraphics[width=0.315\textwidth] {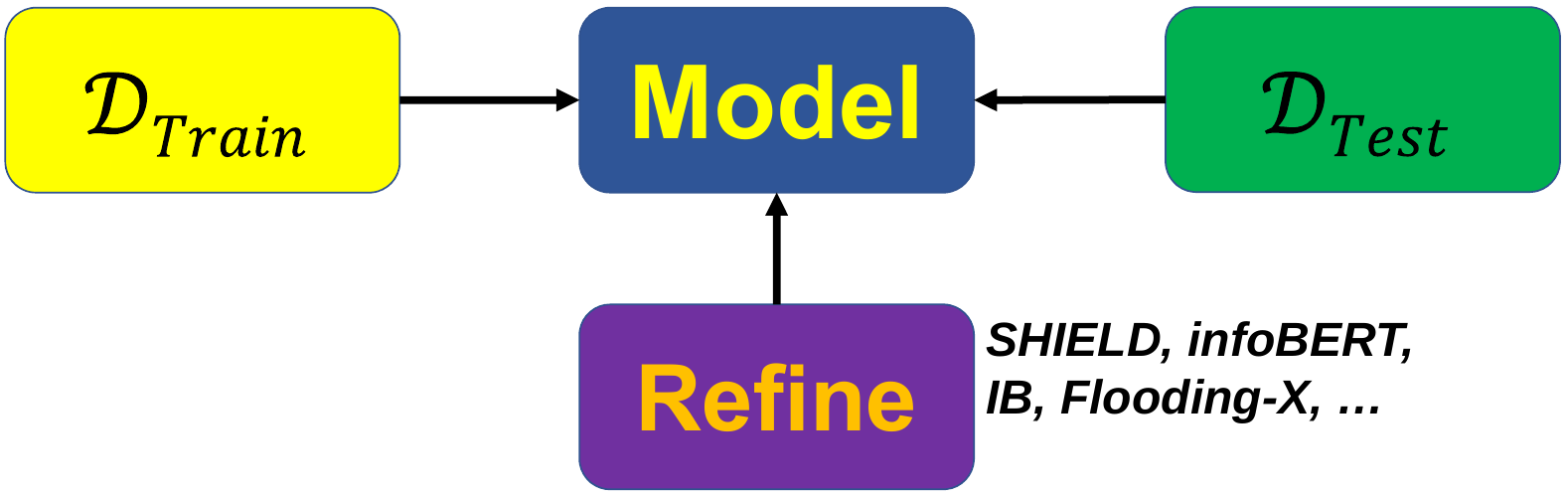}}
~~~\subfigure[randomized smoothing]{\includegraphics[width=0.315\textwidth] {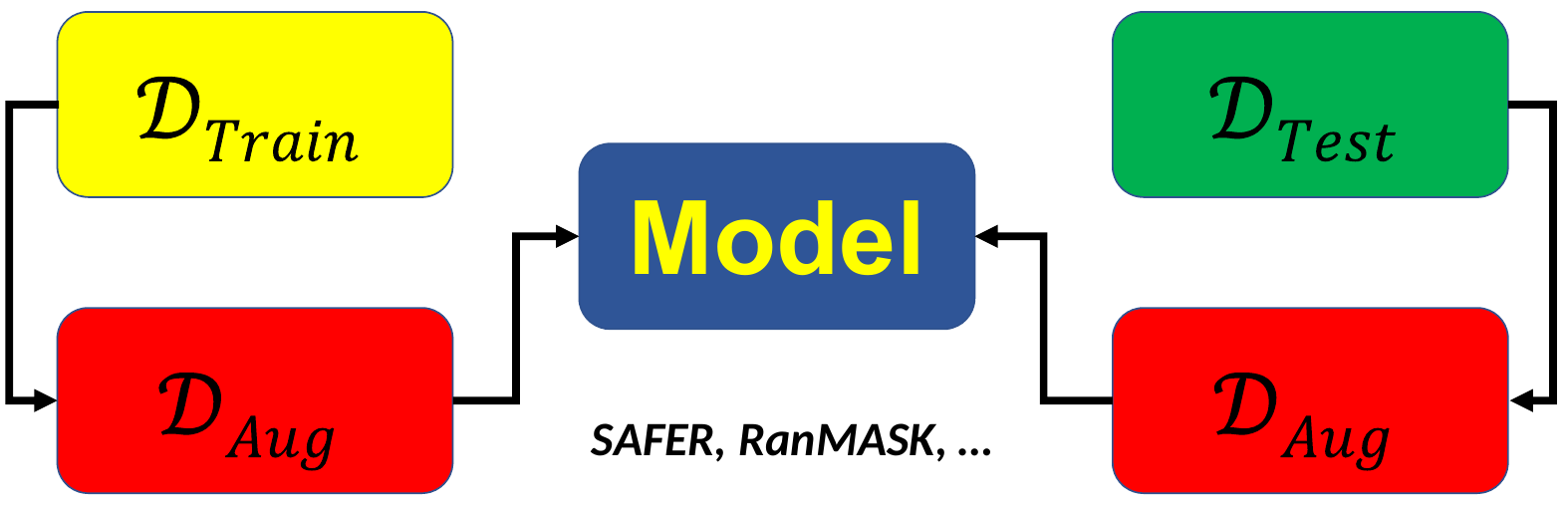}}
	\caption{Comparison of existing adversarial defensing methods.}
	\label{fig:existingDefenses}
\end{figure*}

Text adversarial defenses, in contrast to attacks, aim to form a resilient model capable of maintaining high accuracy on both clean (original) and polluted (adversarial) samples. To mitigate the adverse impact of adversarial attacks, defense methods are typically categorized into three strategies: data augmentation, model adaptation, and randomized smoothing ({as shown in Fig.~\ref{fig:existingDefenses}}).

\noindent \paragraph{\textbf{Data augmentation}} This approach involves strategically augmenting original samples to generate noisy variants, which are then simultaneously utilized to fine-tune the victim model, enhancing its robustness. Importantly, the noise introduced during augmentation typically differs from the tactics employed in attacks (as a black-box manner).  
Specifically, A2T~\cite{yoo-qi-2021-towards-improving} generates noisy variants by employing a gradient-based method to identify crucial words, iteratively substituting them with synonyms using the DistilBERT similarity. 
FreeLB~\cite{Zhu2020FreeLB} and its variants, such as FreeLB++\cite{li-etal-2021-searching}, impose norm-bounded noise on the embeddings of input sentences to produce variants. 
ADFAR\cite{bao-etal-2021-defending} applies frequency-aware randomization to both original and augmented samples (generated through other attacking methods) to create a randomized adversarial set, which is then combined with original samples for model training. 
\textbf{RMLM}~\cite{wang-etal-2023-rmlm} introduces a synonym-based transformation to randomly corrupt input samples (which could be adversarial) before employing an MLM-based defender to reconstruct denoised inputs. This approach generates abundant samples for adversarial training.
A similar approach (\textbf{Adv-Purification}) is presented in~\cite{li-etal-2023-text}. This method injects noise to training samples via masking input texts and subsequently reconstructs masked texts using MLM, forming part of a multiple-run purification process. 
\textbf{GenerAT}~\cite{zhao-mao-2023-generative} randomly masks inputs while simultaneously injecting model gradients to generate perturbed tokens before filling those masked tokens. These generated adversarial variants are then employed to fine-tune the defense model. 
\textbf{MVP}~\cite{raman-etal-2023-model} incorporates a prompt template containing \texttt{[MASK]} token(s) into the input, and performs model prediction by filling the \texttt{[MASK]} token(s) (indicating the labels). A similar approach is adopted by \textbf{RobustSentEmbed}~\cite{rafiei-asl-etal-2024-robustsentembed}, where augmented samples are generated for model fine-tuning.

\noindent \paragraph{\textbf{Model adaptation}} This strategy, without generating noisy variants, refers to enhancing the victim model architecture and/or training losses. For example, \textbf{Infobert}~\cite{wang2021infobert} refines the model by introducing an Information Bottleneck regularizer to suppress noisy information between inputs and latent representations, along with an Anchored Feature regularizer to strengthen the correlation between local and global features. 
Similarly, \textbf{IB}~\cite{zhang-etal-2022-improving} inserts an additional information bottleneck layer between the output layer and the encoder to improve the robustness of the extracted representation. 
\textbf{SHIELD}~\cite{le-etal-2022-shield} modifies the last layer of the victim model, formulating as an ensemble of multiple-expert predictors with random weights. 
\textbf{Flooding-x}~\cite{liu2022flooding} adopts the gradient consistency criterion as a threshold to monitor the training loss and further introduces an early-stop regularization technique to prevent overfitting of training samples. 
\textbf{SIWCon}~\cite{zhan-etal-2023-similarizing} introduces a contrastive learning-based loss, aiming to ensure that less important input words/tokens have a comparable influence on model predictions as their more important counterparts. 
\textbf{ATINTER}~\cite{gupta-etal-2023-dont} integrates an additional encoder-decoder module to rewrite adversarial inputs, eliminating adversarial perturbations before the model inference. 
\textbf{ROIC-DM}~\cite{yuan2024roic} introduces a robust text inference and classification model that leverages diffusion-based architectures with integrated denoising stages, enhancing resistance to adversarial attacks without compromising performance. This model surpasses traditional language models in robustness and, by incorporating advisory components, achieves comparable or superior performance, as validated through extensive testing across multiple datasets.
\textbf{DiffuseDef}~\cite{li2024diffusedef} is a novel adversarial defense approach for language models that enhances robustness by integrating a diffusion layer used as a denoiser between the encoder and classifier, combining adversarial training, iterative denoising, and ensembling techniques to significantly outperform existing methods in resisting adversarial attacks.
And \textbf{FAT}~\cite{yang2024fast}enhances the adversarial robustness of natural language processing models by using a single-step gradient ascent to generate adversarial examples in the embedding space, capitalizing on the consistency of perturbations over training epochs without the need for preset linguistic knowledge.

\noindent \paragraph{\textbf{Randomized smoothing}} This approach employs an ensemble-based approach to enhance the model vulnerability to adversarial attacks. 
\textbf{SAFER}~\cite{ye-etal-2020-safer}, for instance, constructs stochastic input ensembles and utilizes statistical properties of ensembles for classifying testing samples. 
In \textbf{RanMASK}~\cite{zeng-etal-2023-certified}, a few input tokens are randomly substituted using \texttt{[MASK]} for fine-tuning, and testing samples are also masked at different locations to generate multiple masked versions. The final prediction is determined by a majority vote from the ensemble of these masked versions. 
\textbf{RSMI}~\cite{moon-etal-2023-randomized} is a two-stage framework that utilizes randomized smoothing and masked inference. In the first stage, stochastic smoothing is employed to establish a smooth classifier. In the second stage, tokens with significant loss gradients are chosen for masking using multiple Monte-Carlo sampling. The final prediction is obtained by averaging predictions from all these masked samples.

\section{Proposed method}\label{sec:proposedmethod}
\begin{figure}[!th]
\centering
\includegraphics[width=\linewidth]{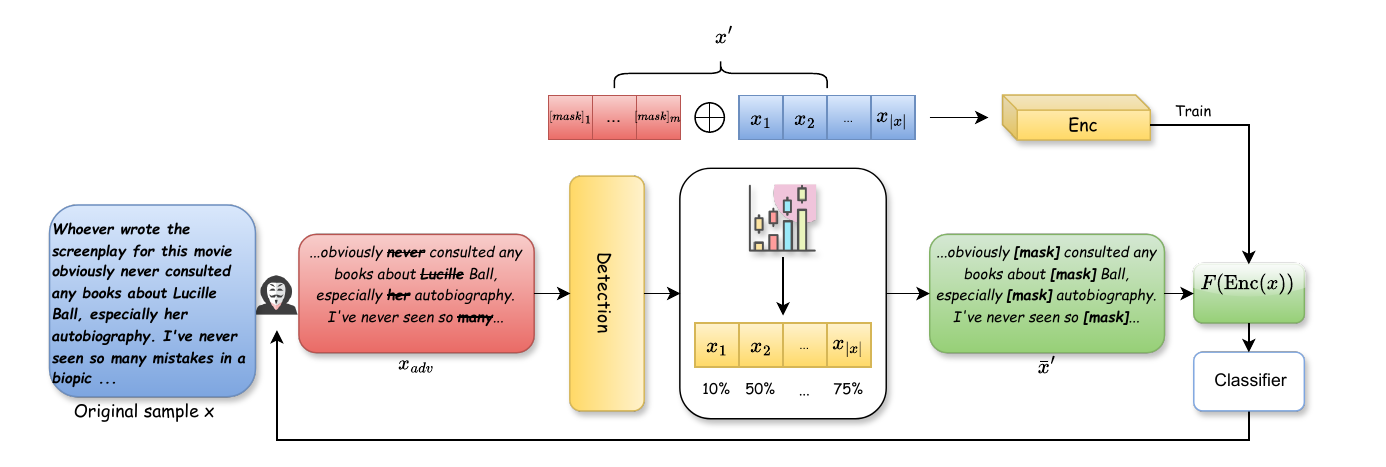}
\caption{The workflow of our proposed method, that preserves the encoder architecture and loss function as the vanilla model. Its distinctiveness lies in integrating  \texttt{[MASK]} tokens into input sequences during both training and inference stages.}
\label{fig:framework}
\end{figure}

This section introduces a simple yet effective algorithm designed to enhance the model resilience against adversarial attacks, termed \textbf{D}efensive \textbf{D}ual \textbf{M}asking (\textbf{\ourmethod}). The proposed method is characterized by strategically injecting \texttt{[MASK]} tokens into input sequences during both training and inference stages. {The workflow of the proposed \ourmethod is shown in Fig.~\ref{fig:framework}}.

\subsection{Defensive Dual Masking}
The proposed method involves two primary stages. In the \textit{training} stage, \ourmethod follows the standard fine-tuning process, utilizing the identical network architecture and training loss as the vanilla model. However, it introduces a unique step of randomly inserting \texttt{[MASK]} tokens into input sequences. In the \textit{inference} stage, our method substitutes potentially adversarial tokens with \texttt{[MASK]} before forwarding the sequence for prediction. Compared to existing defense methods, our approach does not alter the vanilla model (neither its architecture nor training loss) nor necessitates complex strategies for generating noisy variants of original samples. Additionally, our method does not require masking input sequences multiple times for an ensemble-based learning approach, as in the randomized smoothing approach.

Specifically, considering the tokenized (clean) input sequence $\bm{x}$ (\ie, $\bm{x} =$ \texttt{[CLS]} $\bm{x}_1$$\cdots$$\bm{x}_{|\bm{x}|}$ \texttt{[SEP]}), where $\bm{x}_i$ represents the $i$-th token from $\bm{x}$.
In the context of text classification, the goal is to optimize an encoder $\operatorname{Enc}(\cdot)$ and a Multilayer Perceptron (MLP) layer $\mathcal{F}(\cdot)$ to map $\bm{x}$ to a desired label $y$, \ie, $\mathcal{F}(\operatorname{Enc}(\bm{x})) \mapsto y$. Furthermore, let $b_M$ be the predefined masking budget (or the fraction of masked tokens).

During \textit{training}, \ourmethod injects $M$ consecutive masks after \texttt{[CLS]} within $\bm{x}$ to create a masked sequence, denoted as
\[
\bm{x}^\prime = \texttt{[CLS]}~\texttt{[MASK]}_{1}~\cdots~\texttt{[MASK]}_{M}~\bm{x}_{1}~\cdots~\bm{x}_{|\bm{x}|}~\texttt{[SEP]},
\]
where $M$ is determined as $\left \lceil |\bm{x}| \ast b_M \right \rceil$. Subsequently, only $\bm{x}^\prime$ (instead of $\bm{x}$) is utilized for training, and a standard (Cross-Entropy) loss function, denoted as $\mathcal{L}(F(\operatorname{Enc}(\bm{x}^\prime)), y)$, is utilized. 

During \textit{inference}, when presented with an unseen sequence $\bar{\bm{x}}$, our method initially computes a potentially adversarial score for each input token. Subsequently, in a descending manner, tokens with higher scores from $\bar{\bm{x}}$ (say $\bar{\bm{x}}_{i}$ and $\bar{\bm{x}}_{j}$) are successively replaced by \texttt{[MASK]} until the desired number $M$ of masked tokens is reached. This results in a modified sequence:
\[
\bar{\bm{x}}^\prime = \texttt{[CLS]}~\bar{\bm{x}}_{1}~\cdots~\bar{\bm{x}}_{i-1}~\texttt{[MASK]}~\bar{\bm{x}}_{i+1}~\cdots~\bar{\bm{x}}_{j-1}~\texttt{[MASK]}~\bar{\bm{x}}_{j+1}~\cdots~\bar{\bm{x}}_{|\bar{\bm{x}}|}~\texttt{[SEP]}. 
\]
The label of $\bar{\bm{x}}$ accordingly is determined by $\mathcal{F}(\operatorname{Enc}(\bar{\bm{x}}^\prime))$. 
Notably, when either inserting or substituting \texttt{[MASK]} tokens in \ourmethod, we set the position embeddings of \texttt{[MASK]} as zero to minimize the positional impact, while preserving the relevant token and token type embeddings.

\subsection{Analysis on \ourmethod}\label{sec:analysis}
Our approach leverages \texttt{[MASK]} tokens during both the training and inference phases, each with distinct objectives. During training, \texttt{[MASK]} tokens are inserted at the beginning of samples to introduce perturbations that deviate from the original dataset, effectively acting as a form of noise. This use of \texttt{[MASK]} as a \textit{placeholder} helps the model learn to generalize by exposing it to incomplete or partially obscured data, thus enhancing its robustness against unseen or adversarial inputs. In the inference phase, \texttt{[MASK]} tokens are strategically employed to replace potentially adversarial tokens, enabling the model to reduce the influence of adversarial perturbations while preserving the integrity of the underlying context. {The subsequent analysis demonstrates the advantages of replacing adversarial tokens with \texttt{[MASK]}, highlighting its contribution to improved model robustness.}

Let $\vc{a}$, $\vc{r}$, $\mat{S}$, and $\vc{m}$ represent the victim token (being attacked), the replaced token (after the attack), the remaining unchanged tokens, and the \texttt{[MASK]} token, respectively\footnote{Notation: lowercase letters denote single values; bold lowercase/uppercase letters, \eg, $\vc{a}$/$\mat{A}$, represent vectors/matrices}, and the hidden dimension is $d$. Unchanged tokens, \ie, tokens that are not subjected to adversarial attacks, can be collapsed into a single contracted point. The rationale behind is rooted in the attention mechanism of a Transformer model. Consider three tokens, $\vc{x}$, $\vc{s}$, and $\vc{p}$, with their corresponding projections resulting from the linear transformations applied within the attention mechanism:
\[
\vc{x}_i = \vc{x} \mat{W}_i,\quad i=1,2,3,
\]
with similar projections (\ie, $\vc{s}_{(1,2,3)}$ and $\vc{p}_{(1,2,3)}$) for the other two tokens. The reconstructed token $\tilde{\vc{x}}$ after the attention mechanism then is expressed as:
\[
\tilde{\vc{x}} = \frac{\exp(\vc{x}_1 \vc{x}_2^\top) \vc{x}_3 + \exp(\vc{x}_1 \vc{s}_2^\top) \vc{s}_3 + \exp(\vc{x}_1 \vc{p}_2^\top) \vc{p}_3}{\exp(\vc{x}_1 \vc{x}_2^\top) + \exp(\vc{x}_1 \vc{s}_2^\top) + \exp(\vc{x}_1 \vc{p}_2^\top)} 
= \frac{\vc{x}_3^s + w' \vc{p}_3}{1 + w'},
\]
where 
\[
\vc{x}_3^s = \frac{\exp(\vc{x}_1 \vc{x}_2^\top) \vc{x}_3 + \exp(\vc{x}_1 \vc{s}_2^\top) \vc{s}_3}{\exp(\vc{x}_1 \vc{x}_2^\top) + \exp(\vc{x}_1 \vc{s}_2^\top)}, \quad w' = \frac{\exp(\vc{x}_1 \vc{p}_2^\top)}{\exp(\vc{x}_1 \vc{x}_2^\top) + \exp(\vc{x}_1 \vc{s}_2^\top)}.
\]
That is, $\vc{x}_3^s$ is independent of $\vc{p}$, and can be regarded as a contraction of $\vc{x}$ and $\vc{s}$. Due to the flexibility of $\mat{W}_2$, we have $w' \in \mathbb{R}^+$, \ie, $w'$ can take any non-negative real value. {Applying these observations to the tokens in $\mat{S}$ leads to a contracted single point $\vc{s}$, whose position is determined by the model parameters $\mat{W}_i$.}

We continue by analyzing the reconstruction of the \texttt{[CLS]} token as the final input representation. The original reconstructed \texttt{[CLS]} token, denoted as $\tilde{\vc x}_a$, can be computed using the vectors $\vc s$ and $\vc a$. However, $\tilde{\vc x}_a$ is perturbed and replaced by $\tilde{\vc x}_r$, when $\vc a$ is substituted with $\vc r$. In the proposed \ourmethod, we further introduce the \texttt{[MASK]} token, represented as $\vc m$. Ideally, the reconstructed \texttt{[CLS]} token should now be derived solely from $\vc s$ and $\vc m$, denoted as $\tilde{\vc x}_m$.  
Nevertheless, if the perturbation token $\vc r$ is retained, the resulting reconstructed \texttt{[CLS]} lies within the convex hull of $\vc s$, $\vc m$, and $\vc r$, which follows the MI4D process~\cite{hu-etal-2023-mask}. The geometric relationships between these tokens are illustrated in Fig.~\ref{fig:tokengeometry}. Here, $\vc s$ represents the compressed token from the intact token set $\mat S$ (the remaining unchanged tokens). All tokens are assumed to lie on a manifold embedded in $\Real^d$, represented as a smooth surface. 
Due to the attention mechanism, the reconstructed \texttt{[CLS]} token must reside within the convex hull formed by the relevant tokens. For example, $\tilde{\vc x}_a$ lies within $\mathrm{conv}\{\vc s, \vc a\}$, where $\mathrm{conv}\{\vc s, \vc a\}$ denotes the convex hull between $\vc s$ and $\vc a$, geometrically a straight line connecting $\vc s$ and $\vc a$. Similarly, this applies to other reconstructions, such as $\tilde{\vc x}_m$ and $\tilde{\vc x}_r$. Additionally, we have $\tilde{\vc x}_{mr} \in \mathrm{conv}\{\vc s, \vc m, \vc r\}$, where $\mathrm{conv}\{\vc s, \vc m, \vc r\}$ forms a triangle enclosed by $\vc s$, $\vc m$, and $\vc r$. 
Given the model's high complexity, precisely locating the reconstructed \texttt{[CLS]} token is challenging. Therefore, we assume uniformity in the distribution of the reconstructed token within the convex hull of the relevant tokens, which we formalize in the following assumption:
\begin{assumption}\label{a:uniformity}
The reconstructed \texttt{[CLS]} token is uniformly distributed within the convex hull formed by the relevant tokens.
\end{assumption}

\begin{figure}[htp]
    \centering
\tikzset{every picture/.style={line width=0.75pt}} 

\begin{tikzpicture}[x=0.75pt,y=0.75pt,yscale=-1,xscale=1]

\draw    (109,135.75) .. controls (127.5,96.25) and (266.5,7.75) .. (338,60) .. controls (409.5,112.25) and (396.5,108.25) .. (439,104.75) ;
\draw [color={rgb, 255:red, 155; green, 155; blue, 155 }  ,draw opacity=1 ]   (321.5,58.25) .. controls (337,66.25) and (340.5,70.75) .. (358,84.25) ;
\draw [color={rgb, 255:red, 155; green, 155; blue, 155 }  ,draw opacity=1 ]   (323.5,66.25) .. controls (339,74.25) and (336.5,74.25) .. (354,87.75) ;
\draw [color={rgb, 255:red, 155; green, 155; blue, 155 }  ,draw opacity=1 ]   (401,110.25) .. controls (409,114.75) and (409,113.25) .. (425.5,112.25) ;
\draw [color={rgb, 255:red, 155; green, 155; blue, 155 }  ,draw opacity=1 ]   (403,114.25) .. controls (411,118.75) and (411,117.25) .. (427.5,116.25) ;

\draw  [->,line width=1pt] (142,128) -- (275,54) ;

\draw  [->,line width=1pt] (142,128) -- (302,98.37) ;

\draw  [->,line width=1pt] (142,128) -- (391,132.46) ;

\draw  (310,98) -- (394,129) ;

\draw [fill=black] (213,88.5) circle [x radius= 2, y radius= 2] ;
\draw (200.5,68.9) node [anchor=north west][inner sep=0.75pt]    {$\tilde{\vc x}_a$};
\draw [dash pattern={on 2pt off 1pt}] (213,88.5) -- (240,110) ;
\draw [dash pattern={on 2pt off 1pt}] (213,88.5) -- (248,130.5) ;
\draw [dash pattern={on 2pt off 1pt}] (213,88.5) -- (271,120) ;

\draw [fill=black] (240,110) circle [x radius= 2, y radius= 2] ;
\draw (240,89) node [anchor=north west][inner sep=0.75pt]    {$\tilde{\vc x}_m$};

\draw [fill=black] (248,130.5) circle [x radius= 2, y radius= 2] ;
\draw (241,134) node [anchor=north west][inner sep=0.75pt]    {$\tilde{\vc x}_r$};

\draw [fill=black] (271,120) circle [x radius= 2, y radius= 2] ;
\draw (274,111) node [anchor=north west][inner sep=0.75pt]    {$\tilde{\vc x}_{mr}$};

\draw   (137,128) circle [x radius= 4, y radius= 4] ;
\draw (130,114) node [anchor=north west][inner sep=0.75pt]    {$\vc s$};
\draw   (278,52) circle [x radius= 4, y radius= 4] ;
\draw (285.5,49) node [anchor=north west][inner sep=0.75pt]    {$\vc a$};
\draw   (306,98) circle [x radius= 4, y radius= 4] ;
\draw (311.5,88) node [anchor=north west][inner sep=0.75pt]    {$\vc m$};
\draw   (395.5,132.5) circle [x radius= 4, y radius= 4] ;
\draw (402.5,125) node [anchor=north west][inner sep=0.75pt]    {$\vc r$};

\end{tikzpicture}
    \caption{The token geometry where $\vc{a}$, $\vc{r}$, $\vc{s}$, and $\vc{m}$ represent the victim token, replaced token, compressed unchanged tokens, and \texttt{[MASK]} token, respectively. $\tilde{\vc x}_a$, $\tilde{\vc x}_m$, $\tilde{\vc x}_r$, and $\tilde{\vc x}_{mr}$ denote the reconstructed \texttt{[CLS]} token using combinations of $\vc{a}$, $\vc{m}$, $\vc{r}$, and $\vc{s}$.}
    \label{fig:tokengeometry}
\end{figure}

\noindent For example, $\tilde{\vc x}_m$ is then uniformly distributed within $\mathrm{conv}\{\vc s, \vc m\}$. Now we can establish the expected distance between a pair of reconstructed \texttt{[CLS]} by the follow Lemmas. 
\begin{lem}\label{lem:expdist2vecs}
Let two vectors $\vc a$, $\vc b\in\Real^d$. $\tilde{\vc a}$ uniformly distributed between the origin $\vc o$ and $\vc a$, and similarly $\tilde{\vc b}$ uniformly distributed between  $\vc o$ and $\vc b$ independent of $\tilde{\vc a}$. Then 
\[
\mathbb E(\|\tilde{\vc a}-\tilde{\vc b}\|^2) = \frac13\|\vc a\|^2+\frac13\|\vc b\|^2-\frac12\vc a^\top\vc b,
\]
where $\|\vc x\|$ is the $\ell_2$ norm of vector $\vc x$ and $\mathbb E$ is the expectation. 
\end{lem}
\begin{proof}
We rewrite the expectation as
\[
\mathbb E(\|\tilde{\vc a}-\tilde{\vc b}\|^2) = \mathbb E(\|\alpha\vc a-\beta \vc b\|^2),
\]
where $\alpha$ and $\beta$ are two independent random variables uniformly distributed in $[0,1]$, \ie, $\alpha$, $\beta\sim\mathcal U(0,1)$. Then we have 
\begin{align*}
\mathbb E(\|\alpha\vc a-\beta \vc b\|^2) &= \mathbb E(\alpha^2\|\vc a\|^2+\beta^2\|\vc b\|^2-2\alpha\beta\vc a^\top\vc b)\\
&=\mathbb E(\alpha^2)\|\vc a\|^2 + \mathbb E(\beta^2)\|\vc b\|^2 - 2\mathbb E(\alpha\beta)\vc a^\top\vc b\\
&= \frac13\|\vc a\|^2+\frac13\|\vc b\|^2-\frac12\vc a^\top\vc b.
\end{align*}
In the last equality, we used the second momentum of uniform distribution, \ie, $\mathbb E(\beta^2)=\frac13$, and expectation of a product of two independent random variables, \ie, $\mathbb E(\alpha\beta)=\mathbb E(\alpha)\mathbb E(\beta)$. 
\end{proof}
\noindent We further extend the above to the case with three vectors. 
\begin{lem}\label{lem:expdist3vecs}
Let three vectors $\vc a$, $\vc b$, $\vc c\in\Real^d$. $\tilde{\vc a}$ uniformly distributed between the origin $\vc o$ and $\vc a$, and $\tilde{\vc m}$ uniformly distributed within $\mathrm{conv}\{\vc o,\vc b, \vc c\}$ independent of $\tilde{\vc a}$. Then 
\[
\mathbb E(\|\tilde{\vc a}-\tilde{\vc m}\|^2) = \frac13\|\vc a\|^2+\frac19\|\vc b\|^2+\frac19\|\vc c\|^2-\frac14\vc a^\top\vc b-\frac14\vc a^\top\vc c+\frac19\vc b^\top\vc c.
\]
\end{lem}
\begin{proof}
Let $x,y,z\sim\mathcal U(0,1)$ independently. We first construct $\tilde{\vc m}$ from $\vc b$ and $\vc c$ as 
\[
\tilde{\vc m} = z(y\vc b+(1-y)\vc c). 
\]
It is easily to see that $\tilde{\vc m} \subseteq \mathrm{conv}\{\vc o,\vc b, \vc c\}$ as the coefficient for $\vc o$ is $1-z(y+(1-y))=1-z\in[0,1]$. By following a similar approach as outlined in the proof of Lemma \ref{lem:expdist2vecs}, we can rewrite the expectation as follows:
\[
\mathbb E(\|\tilde{\vc a}-\tilde{\vc m}\|^2) = \mathbb E(\|x\vc a- yz\vc b-z(1-y)\vc c\|^2),
\]
and we further derive 
\begin{align*}
\mathbb E(\|x\vc a- yz\vc b-z(1-y)\vc c\|^2) &= \mathbb E(x^2\|\vc a\|^2+y^2z^2\|\vc b\|^2+(1-y)^2z^2\|\vc c\|^2\\
&\qquad -2xyz\vc a^\top\vc b-2xz(1-y)\vc a^\top\vc c+2y(1-y)z^2\vc b^\top\vc c)\\
&=\mathbb E(x^2)\|\vc a\|^2 + \mathbb E(y^2z^2)\|\vc b\|^2 + \mathbb E((1-y)^2z^2)\|\vc c\|^2\\
&\qquad -2\mathbb E(xyz)\vc a^\top\vc b-2\mathbb E(xz(1-y))\vc a^\top\vc c+2\mathbb E(y(1-y)z^2)\vc b^\top\vc c.
\end{align*}
Note that $(1-y)\sim\mathcal U(0,1)$ and $\mathbb E(y(1-y))=\mathbb E(y-y^2)=\frac16$. By using independence of $x$, $y$ and $z$, we have $\mathbb E(xyz)=\frac18$, $\mathbb E(z^2y^2)=\mathbb E(z^2)\mathbb E(y^2)=\frac19$ and $\mathbb E(y(1-y)z^2)=\frac1{18}$. Similar results for switching variables. Substituting these values into the above gives the claimed result.
\end{proof}

\begin{thm}[Success condition for \ourmethod]\label{thm:ddmConditions}
Let $\vc v_a$, $\vc v_m$ and $\vc v_r$ be the vectors of $\vc a$, $\vc m$ and $\vc r$ rooted at $\vc s$ respectively, $l_{am}$, $l_{ar}$ and $l_{mr}$ be cosine similarities between $\vc v_a$ and $\vc v_m$, $\vc v_a$ and $\vc v_r$, and $\vc v_m$ and $\vc v_r$, for example $l_{am}=\frac{\vc v_a^\top \vc v_m}{\|\vc v_a\|\|\vc v_m\|}$. Assuming $\tilde{\vc x}_a$, $\tilde{\vc x}_m$, $\tilde{\vc x}_r$ and $\tilde{\vc x}_{mr}$ are uniformly distributed in their corresponding convex hulls (Assumption~\ref{a:uniformity}) and $l_{mr}>0$, then we have 
\begin{equation}\label{e:ineqinexp}
\mathbb E(\|\tilde{\vc x}_a-\tilde{\vc x}_m\|^2)\le \mathbb E(\|\tilde{\vc x}_a-\tilde{\vc x}_{mr}\|^2) \le \mathbb E(\|\tilde{\vc x}_a-\tilde{\vc x}_r\|^2),
\end{equation}
when the following condition is satisfied
\begin{equation}\label{e:condition}
\|\vc v_r\|\ge\max\left\{\boldsymbol 1(\Delta_1\ge0)\frac98(\|\vc v_a(l_{ar}+4\sqrt{\Delta_1}),\boldsymbol 1(\Delta_2\ge0)\frac34(\|\vc v_a(l_{ar}+2\sqrt{\Delta_2}),\frac{\|\vc v_m\|}{l_{mr}}\right\},
\end{equation}
where 
\begin{subequations}
\begin{empheq}[left=\empheqlbrace]{align}
&\Delta_1 = (\frac29\|\vc v_m\|-\frac{\|\vc v_a\|}4l_{am})^2-\frac{\|\vc v_a\|^2}{16}(l_{am}^2-l_{ar}^2), \\
&\Delta_2 = (\frac23\|\vc v_m\|-\frac{\|\vc v_a\|}2l_{am})^2-\frac{\|\vc v_a\|^2}{4}(l_{am}^2-l_{ar}^2), 
\end{empheq}
\end{subequations}
and $\boldsymbol 1(e)$ is an indicator function returning 1 when $e$ is true and 0 otherwise. 
\end{thm}
\begin{proof}
By setting $\vc s$ as the origin, and defining $\vc a = \vc v_a$, $\vc b = \vc v_m$, and $\vc c = \vc v_r$, we can apply Lemma~\ref{lem:expdist2vecs} and Lemma~\ref{lem:expdist3vecs} to derive all the expectations in Eq.~\eqref{e:ineqinexp}. Consequently, we obtain the following result
\begin{subequations}
\label{e:jointineq}
\begin{empheq}[left=\empheqlbrace, right=\empheqrbrace]{align}
&\frac13\|\vc b\|^2-\frac12\vc a^\top\vc b \le \frac13\|\vc c\|^2-\frac12\vc a^\top\vc c \label{e:jointineq1} \\
&\frac13\|\vc b\|^2-\frac12\vc a^\top\vc b \le \frac19\|\vc b\|^2+\frac19\|\vc c\|^2-\frac14\vc a^\top\vc b-\frac14\vc a^\top\vc c+\frac19\vc b^\top\vc c \label{e:jointineq2}      
\end{empheq}
\end{subequations}
if Eq.~\eqref{e:ineqinexp} holds. Note that the common terms related to $\|\vc a\|$ in Eq.~\eqref{e:jointineq} are omitted, as they do not affect the inequalities. 

\noindent We first prove that if inequalities in Eq.~\eqref{e:jointineq} hold, then Eq. \eqref{e:ineqinexp} holds naturally, \ie, 
\begin{equation}\label{e:latterineq}
\frac13\|\vc c\|^2+\frac13\|\vc b\|^2-\frac12\vc a^\top\vc c \ge \frac19\|\vc b\|^2+\frac19\|\vc c\|^2-\frac14\vc a^\top\vc b-\frac14\vc a^\top\vc c+\frac19\vc b^\top\vc c.   
\end{equation}
\noindent Assuming the above Eq.~\eqref{e:latterineq} is false, then we have 
\begin{subequations}
\label{e:jointineq2}
\begin{empheq}[left=\empheqlbrace, right=\empheqrbrace]{align}
&\frac29\|\vc c\|^2-\frac14\vc a^\top\vc c \le \frac19\|\vc b\|^2-\frac14\vc a^\top\vc b+\frac19\vc b^\top\vc c \label{e:jointineq21} \\
&\frac29\|\vc b\|^2-\frac14\vc a^\top\vc b \le \frac19\|\vc c\|^2-\frac14\vc a^\top\vc c+\frac19\vc b^\top\vc c  \label{e:jointineq22}
\end{empheq}
\end{subequations}
where Eq.~\eqref{e:jointineq21} is from the assumed contradiction, and Eq.~\eqref{e:jointineq22} is from the second inequality of Eq.~\eqref{e:jointineq}. 
Adding both sides gives $\frac19\|\vc c\|^2+\frac19\|\vc b\|^2<\frac29\vc b^\top\vc c$ leading to $\|\vc c - \vc b\|^2<0$, which is impossible. Therefore Eq.~\eqref{e:latterineq} must be true and hence we prove the equivalency between Eq.~\eqref{e:ineqinexp} and Eq.~\eqref{e:jointineq}.
\begin{figure}
    \centering
\begin{tikzpicture}[domain=-2:3]
  \draw[->] (-0.2,0) -- (4,0) node[right] {$x$};
  \draw[->] (0,-0.2) -- (0,3.2) node[above] {$f_0(x)/f_e(x)$};
  \draw[very thin,color=gray] (-2.1,-0.32) grid (3.2,3.2);

\draw[dash pattern={on 2pt off 1pt},color=black,domain=-2:3.6]    plot(\x, {\x*\x/9-0.16*\x}) node[right] {$f_e(x)=\frac19x^2 - 0.16x$};
\draw[color=black,line width=0.75pt]    plot(\x, {\x*\x/9-0.24*\x})           node[right] {$f_0(x)=\frac19x^2 - 0.24x$};

\draw[dash pattern={on 2pt off 1pt},color=blue,domain=-2:2.3] plot(\x, {\x*\x/3+0.2*\x}) node[right] {$f_e(x)=\frac13x^2 + 0.2x$};
\draw[color=blue,domain=-2:2.3,line width=0.75pt]    plot(\x, {\x*\x/3+0.48*\x}) node[right] {$f_0(x)=\frac13x^2 + 0.48x$};
\end{tikzpicture}
    \caption{$f_0(x)$ and $f_e(x)$ function values.}
    \label{fig:fx}
\end{figure}
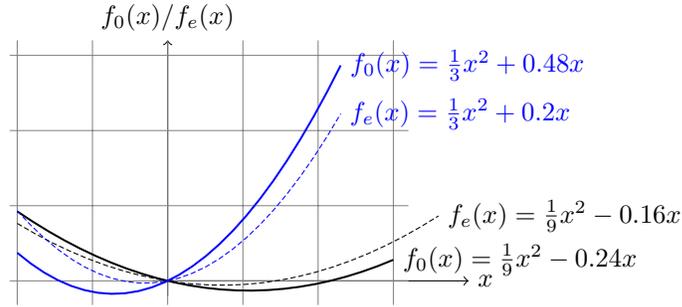

We now focus on Eq.~\eqref{e:jointineq}. Consider a general function in the following form 
\[
f_e(x\mid a,b,l) = ax^2-b(l-e)x, \ (\forall x,a,b>0,\ e\in\Real)
\]
where $a$ and $b$ are fixed constants. Hereafter we omit $|a,b,l$ for simplicity, and some examples of function $f_0(x)$ and $f_e(x)$ are shown in Fig. \ref{fig:fx}. We examine the conditions for $f_e(y)\ge f_0(x)$ for given $x$ and $e$ but varying $y$. Since $a>0$, $f_e(y)\ge f_0(x)$ is always possible and 
\begin{equation}\label{e:fcond}
f_e(y)\ge f_0(x) \Leftrightarrow y\in\left\{
\begin{array}{l}
\Real^+-\frac1{2a}\mathcal B(b(l-e)),\sqrt{\Delta}), \ \Delta\ge0 \\
\Real^+, \ \Delta<0
\end{array},
\right.  
\end{equation}
where $\Delta=4af_0(x)+b^2(l-e)^2$ and $\mathcal B(x,r)=(x-r,x+r)$ is the disc at $x$ with radius $r$ without the boundary. The above is derived by solving $f_e(y)=f_0(x)$ for a fixed $x$. We identify Eq. \eqref{e:jointineq} to $f_e(x\mid \frac13,\frac12\|\vc a\|,l_{am})$ and $f_e(x\mid \frac19,\frac14\|\vc a\|,l_{am})$, for example, Eq. \eqref{e:jointineq1} can be rewritten as 
\[
f_0(\|\vc b\|\mid \frac13,\frac{\|\vc a\|}2,l_{am}) \le f_e(\|\vc c\|\mid\frac13,\frac{\|\vc a\|}2,l_{am}), 
\]
where $e$ is to account for the difference between $l_{am}$ and $l_{ar}$ so that $l_{ar}=l_{am}-e$. By using Eq. \eqref{e:fcond}, we observe that 
\begin{equation}\label{e:eqiv1}
\eqref{e:jointineq1} \Leftrightarrow y\in \Real^+-\frac98\mathcal B(\|\vc v_a\|l_{ar},4\sqrt{\Delta_1}). 
\end{equation}
Similarly we have 
\begin{equation}\label{e:eqiv2}
\frac19\|\vc b\|^2-\frac14\vc a^\top\vc b \le \frac19\|\vc c\|^2-\frac14\vc a^\top\vc c  \Leftrightarrow y\in \Real^+-\frac34\mathcal B(\|\vc v_a\|l_{ar},2\sqrt{\Delta_2}).
\end{equation}
It is important to note that in both Eq.~\eqref{e:eqiv1} and Eq.~\eqref{e:eqiv2}, we omit the case when $\Delta_{i}$, for $i=1,2$, where $y \in \Real^+$. 
It is now evident that the desired condition defines a subset of the intersection of the sets specified in Eq.~\eqref{e:eqiv1} and Eq.~\eqref{e:eqiv2} (appearing on the right-hand sides of the inequalities). By leveraging the fact that $x^2 \leq xyl_{mr}$ when $y \geq \frac{x}{l_{mr}}$ for $l_{mr} > 0$, we deduce that $\|\vc{b}\|^2 \leq \vc{b}^\top \vc{c}$ holds when $\|\vc{v_r}\| \geq \frac{\|\vc{v_m}\|}{l_{mr}}$. This completes our proof.
\end{proof}

\begin{rem}
Theorem \ref{thm:ddmConditions} establishes that, when the condition in Eq.~\eqref{e:condition} is satisfied, the reconstructed \texttt{[CLS]} token, obtained by leveraging only the \texttt{[MASK]} token and residual tokens, is closer to the original reconstructed \texttt{[CLS]} token (before the attack) than any other reconstructed version. This result is formalized in the inequality presented in Eq.~\eqref{e:ineqinexp}. Consequently, this validates the proposed \ourmethod in identifying the potentially adversarial token(s) for masking. 
In cases where multiple tokens are attacked, the analysis remains applicable due to the contraction in convex reconstruction. Specifically, the vectors $\vc a$ and $\vc r$ can be understood as contracted versions of the vector forms of the attacked/victim tokens and the replacement tokens, respectively. Notably, the condition in Eq.~\eqref{e:condition} is relatively mild, primarily stating that the distributions of $\vc a$, $\vc m$, and $\vc r$ are centered around $\vc s$. Although the underlying manifold may influence the token locations, as it directly impacts their spatial configuration, the analysis itself does not depend on any specific manifold geometry.
\end{rem}



%
%

\section{Experiments}\label{sec:experiments}
\subsection{Setup}

{\paragraph{\textbf{Datasets}} To evaluate the effectiveness of \ourmethod, we conducted experiments on highly competitive text classification benchmark datasets: \textbf{AGNews}~\cite{AGNEWS} and \textbf{MR}~\cite{mr}. Table~\ref{tab:datasets} provides statistics on these datasets.}

%
\begin{table}[!th]
    \centering
    \caption{Statistics on employed text classification benchmarks, where \#Train, \#Test, \#Min, \#Max, \#Median, and \#Length represent, respectively, instances in the training and test sets, minimum length, maximum length, median length, and the average length of input text.}\label{tab:datasets}
    \begin{adjustbox}{max width=0.75\textwidth}
    \begin{tabular}{l|cccccc}
        \toprule
        Dataset  & \#Train & \#Test   	&\#Min 	&\#Max 	&\#Median 	&\#Length   \\ \midrule
        AGNEWS   & 120,000 & 7,600 		&8 		&177 	&37 		& 40\\
        MR  	 &  8530   & 1070       &1   	&59 	&20 	  	&22  \\
         \bottomrule
    \end{tabular} 
    \end{adjustbox}
\end{table}

\paragraph{\textbf{Attacking Algorithms}} {To demonstrate that our strategy can handle different adversarial attacks, we employed the well-known TextAttack framework~\cite{morris2020textattack} and utilized four different attack strategies. The details of specific attack methods are as follows:}
\begin{itemize} \setlength{\itemsep}{-0.5pt}
	\item \textbf{TextFooler}~\cite{jin2020bert} introduces word-level perturbations by replacing original words with their synonyms.
	\item \textbf{BERT-Attack}~\cite{li2020bert} applies word-level perturbations by leveraging a pre-trained masked language model to substitute target words.
	\item \textbf{DeepWordBug}~\cite{DeepWordBug} focuses on character-level perturbations, including substitutions, deletions, insertions, and letter swaps within words.
	\item \textbf{TextBugger}~\cite{TextBugger} combines both symbol- and word-level perturbations, utilizing techniques like inserting spaces, replacing words, deleting characters, and swapping adjacent letters to craft adversarial examples.
\end{itemize}

\paragraph{\textbf{Defense Algorithms}} The proposed \ourmethod is evaluated against state-of-the-art methods across three categories, as outlined below: 
\begin{itemize}
    \item \textbf{Data Augmentation}: \textbf{FreeLB++}~\cite{li-etal-2021-searching}, \textbf{RMLM}~\cite{wang-etal-2023-rmlm}, \textbf{Adv-Purification}~\cite{li-etal-2023-text}, \textbf{MVP}~\cite{raman-etal-2023-model}, and \textbf{FAT}~\cite{yang2024fast}.    
    \item \textbf{Model Enhancements}: \textbf{InfoBERT}~\cite{wang2021infobert}, \textbf{Flooding-x}~\cite{liu2022flooding}, \textbf{RobustT}~\cite{zheng-etal-2022-robust},\textbf{ATINTER}~\cite{gupta-etal-2023-dont}, \textbf{SIWCon}~\cite{zhan-etal-2023-similarizing}, \textbf{LLMPM}~\cite{moraffah2024adversarial},and \textbf{ROIC-DM}~\cite{yuan2024roic}.    
    \item \textbf{Random Smoothing}: \textbf{RanMASK}~\cite{zeng-etal-2023-certified}, \textbf{RSMI}~\cite{moon-etal-2023-randomized}, and \textbf{MI4D}~\cite{hu-etal-2023-mask},\textbf{Text-RS}~\cite{zhang2024random}.
\end{itemize}
{Among these methods, \textbf{RanMASK}, \textbf{RMLM}, \textbf{RSMI}, \textbf{Adv-Purification}, \textbf{MVP}, \textbf{RobustT} and \textbf{MI4D}  also utilize the masking approach. All contender methods are reviewed in Section~\ref{sec:existingDefenses}, with their results directly drawn from the respective original papers.}

\paragraph{\textbf{Implementation}} For our main experiments (Section~\ref{sec:mainExperiments}), we employ the BERT-base model~\cite{devlin-etal-2019-bert} as the encoder. Training is conducted using a batch size of 32 sequences, each with a maximum length of 128 tokens. We reserve 10\% of the training set for validation, and early stopping is applied if the validation accuracy does not improve within one epochs or after reaching a maximum of 10 epochs. A dropout rate of 0.1 is applied across all layers. We utilize the Adam optimizer with a learning rate that warms up to $2e^{-5}$ over the first 10,000 steps and then decays linearly to $1e^{-6}$ following a cosine annealing schedule. Gradient clipping is enforced within the range $(-1, 1)$. 
During inference, \ourmethod masks potentially adversarial tokens by employing the Word Frequency-based (FGWS) strategy~\cite{mozes-etal-2021-frequency} to estimate the perturbation probability of each candidate token. Specifically, we calculate the occurrence frequency of each candidate token within the training dataset, and the perturbation probability is estimated as the ratio of the token frequency to the total (training) corpus size. {The masking budget for training and testing is set as 30\%.}
All experiments are repeated five times using different random seeds, with the results averaged to ensure reliability. Finally, all computations are performed on an NVIDIA A100 GPU server.

\paragraph{\textbf{Evaluation Metrics}} We follow the experimental setup outlined in~\cite{wang2021infobert, zhang-etal-2022-improving, zeng-etal-2023-certified}. That is, we select 1,000 samples that were successfully attacked, originally drawn at random from the testing dataset, to evaluate the model robustness against adversarial attacks. 
The following three key metrics are employed: (1) \textbf{CLA\%}, which represents the classification accuracy of the model on the original, clean data; (2) \textbf{CAA\%}, denoting the classification accuracy under specific adversarial attacks. \textbf{A higher CAA\% reflects better defense performance}; and (3) \textbf{SUCC\%}, which measures the success rate of adversarial attacks, defined as the proportion of examples successfully misclassified out of the total attack attempts. \textbf{A lower SUCC\% indicates greater robustness of the model}.

\subsection{Main results}\label{sec:mainExperiments}
\begin{table*}[!th]
\caption{Adversarial defense performance comparison between \ourmethod and existing methods. Superscripts $(1)$, $(2)$, and $(3)$ indicate the use of data augmentation, model enhancement, and random smoothing methods, respectively. {Methods employing the masking strategy are denoted with $^*$. The best performance is highlighted in \textbf{bold}, while the second-best is \underline{underlined}.} Statistically significant at $p<10^{-3}$ are marked with $\dagger$.}
\label{tab:mainRes}
\resizebox{\linewidth}{!}{
\begin{tabular}{c|l|c|cc|cc|cc|cc}
\hline
\hline
\multirow{2}{*}{\bf Datasets} & \multirow{2}{*}{\bf  Methods} & \multirow{2}{*}{\bf CLA\%} & \multicolumn{2}{c|}{\textbf{TextFooler}} & \multicolumn{2}{c|}{\textbf{BERT-Attack}} & \multicolumn{2}{c|}{\textbf{Deepwordbug}} & \multicolumn{2}{c}{\textbf{TextBugger}}  \\ \cline{4-11} 
&  &  & \emph{\bf CAA\%} & \emph{\bf SUCC\%}   & \emph{\bf CAA\%} & \emph{\bf SUCC\%}   & \emph{\bf CAA\%} & \emph{\bf SUCC\%}   & \emph{\bf CAA\%} & \emph{\bf SUCC\%}  \\ \hline

\multirow{17}{*}{\textbf{AGNEWS}}
&Baseline			&$92.8$ &$15.8$ &$83.2$ &$26.7$ &$71.8$ &$33.0$ 	&$65.0$ &$49.2$ &$47.8$   \\
&FreeLB++$^{(1)}$	&$93.3$ &$51.5$ &$46.0$ &$41.8$	&$56.2$ &$55.1$	&$42.1$ &$55.9$	&$41.4$  \\
&RMLM$^{(1)*}$ 		&$94.0$ &$81.0$ &$13.7$ &$48.1$ &$48.7$ &$-$ 	&$-$ 	&$-$ 	&$-$ \\
&Adv-Purification$^{(1)*}$ 
					&$92.0$ &$61.5$ &$27.9$ &$49.7$ &$46.5$ &$-$ 	&$-$ 	&$-$ 	&$-$ \\
&MVP$^{(1)*}$		&$94.5$ &$53.7$ &$39.3$ &$\underline{83.4}$ &$\underline{10.4}$ &$-$ 	&$-$ 	&$69.2$ &$25.4$ \\
&FAT$^{(1)}$		&$95.1$ &$62.3$ &$33.2$ &$48.0$ &$47.4$ &$-$ 	&$-$ 	&$63.6$ &$32.4$ \\

&ROIC-DM$^{(2)}$	&$94.1$	&$78.7$	&$18.4$	&$49.0$	&$47.0$	&$-$	&$-$	&$-$	&$-$   \\
&InfoBERT$^{(2)}$	&$93.2$	&$51.3$	&$45.2$	&$39.9$	&$57.4$	&$53.9$	&$42.4$	&$50.6$	&$45.9$   \\
&ATINTER$^{(2)}$    &$92.6$ &$73.0$ &$21.1$ &$22.9$ &$75.3$ &$21.7$ &$76.6$	&$63.9$	&$32.5$  \\
&Flooding-X$^{(2)}$	&$92.7$ &$68.9$ &$27.0$ &$56.4$ &$40.3$ &$65.3$ &$30.8$ &$70.3$ &$25.5$   \\
&SIWCon$^{(2)}$     &$92.6$ &$19.7$ &$78.7$ &$23.1$ &$75.1$ &$20.6$ &$77.8$	&$52.9$	&$42.9$  \\
&RobustT$^{(2)*}$	&$94.9$ &$28.5$ &$70.0$ &$12.1$ &$87.2$ &$-$    &$-$    &$53.4$ &$43.7$ \\
&LLMPM$^{(2)}$      &$95.1$ &$\underline{81.3}$  &$\underline{13.4}$ &$-$    &$-$    &$-$    &$-$    &$-$    &$-$\\

&RanMASK$^{(3)*}$	&$92.6$ &$37.9$	&$58.7$	&$49.5$	&$46.1$	&$38.4$	&$54.6$	&$45.0$	&$50.9$\\
&RSMI$^{(3)*}$		&$92.9$	&$71.7$	&$22.8$	&$67.5$	&$27.3$ &$\underline{72.8}$	&$\underline{21.6}$	&$69.7$	&$25.0$\\
&MI4D$^{(3)*}$			&$93.0$	&$66.7$	&$29.2$	&$69.7$	&$25.9$	&$62.4$	&$34.0$	&$\underline{73.9}$	&$\underline{21.9}$\\
        
&\cellcolor[gray]{.9}\bf \ourmethod  
        &\cellcolor[gray]{.9}$\textbf{92.8} \dagger$
		&\cellcolor[gray]{.9}$\textbf{82.3} \dagger$	&\cellcolor[gray]{.9}$\textbf{10.6} \dagger$	
		&\cellcolor[gray]{.9}$\textbf{84.2} \dagger$	&\cellcolor[gray]{.9}$\textbf{8.8} \dagger$
		&\cellcolor[gray]{.9}$\textbf{85.8} \dagger$	&\cellcolor[gray]{.9}$\textbf{8.7} \dagger$
		&\cellcolor[gray]{.9}$\textbf{83.3} \dagger$	&\cellcolor[gray]{.9}$\textbf{10.1} \dagger$\\ 
     
\hline

\multirow{10}{*}{\textbf{MR}}  
&Baseline			&$85.7$ &$6.5$ &$92.4$  &$8.7$ &$89.8$  &$19.4$ &$77.4$  &$26.3$ &$69.3$ \\
&FreeLB++$^{(1)}$	&$86.4$ &$10.8$ &$87.5$  &$11.2$ &$87.0$  &$22.3$ &$74.2$  &$31.8$ &$63.2$ \\
        
&ATINTER$^{(2)}$       &$85.6$ &$21.1$ &$74.6$  &$19.3$ &$77.5$  &$30.6$ &$64.3$  &$45.7$ &$47.8$ \\
&Flooding-X$^{(2)}$    &$85.6$ &$21.1$ &$75.4$  &$20.4$ &$76.2$  &$28.9$ &$66.2$  &$31.6$ &$63.1$ \\
&SIWCon$^{(2)}$        &$85.4$ &$30.7$ &$64.2$  &$17.6$ &$79.4$  &$\underline{60.3}$ &$\underline{36.1}$  &$36.8$ &$56.9$ \\
&RanMASK$^{(3)*}$       &$85.1$ &$12.9$ &$84.8$  &$16.8$ &$80.3$  &$19.6$ &$77.0$  &$33.1$ &$61.1$ \\
&RSMI$^{(3)*}$          &$86.1$ &$\underline{47.6}$ &$\underline{44.7}$  &$39.5$ &$54.1$  &$58.1$ &$32.5$  &$\underline{56.4}$ &$\underline{34.5}$ \\
&MI4D$^{(3)*}$            &$85.8$ &$43.7$ &$49.1$  &$\underline{40.8}$ &$\underline{52.4}$  &$55.5$ &$35.3$  &$51.2$ &$40.3$ \\
      
&\cellcolor[gray]{.9}\bf \ourmethod  
        &\cellcolor[gray]{.9}$\textbf{85.7}\dagger$ 
        &\cellcolor[gray]{.9}$\textbf{55.2}\dagger$		&\cellcolor[gray]{.9}$\textbf{37.9}\dagger$  
        &\cellcolor[gray]{.9}$\textbf{62.6}\dagger$		&\cellcolor[gray]{.9}$\textbf{29.9}\dagger$  
        &\cellcolor[gray]{.9}$\textbf{68.7}\dagger$		&\cellcolor[gray]{.9}$\textbf{19.6}\dagger$  
        &\cellcolor[gray]{.9}$\textbf{63.3}\dagger$		&\cellcolor[gray]{.9}$\textbf{26.5}\dagger$ \\

\hline
\end{tabular}
}
\end{table*}
The average results from five trials for adversarial defense performance are summarized in Table~\ref{tab:mainRes}. The following key observations can be made: (1) \textbf{On classification accuracy}. The proposed method does not compromise the classification accuracy on clean testing data (CLA\%). Specifically, it maintains performance on par with regular fine-tuning and does not introduce additional data, as seen in data augmentation techniques. As a result, the CLA\% remains consistent across all datasets. (2) \textbf{In terms of defense accuracy}. Our approach demonstrates superior results compared to existing methods across all datasets. For instance, against the \textbf{TextFooler} attack, our method achieves excellent performance on two datasets, where it achieves a CAA\% of 82.3\% and 55.2\%. For \textbf{DeepWordBug}, \ourmethod outperforms all other methods, achieving the highest reported CAA\% values of 85.8\% and 68.7\% across all datasets, demonstrating its robustness against this attack. {(3) \textbf{Different Masking strategy}. We further explore the utilization of masking strategies across various defense methods, including our own. From the results presented in Table \ref{tab:mainRes}, incorporating masking into Data Augmentation is more effective against word-substitution attacks such as TextFooler and BERT-Attack, while integrating masking into Random Smoothing proves more effective against character-level attacks like DeepWordBug and TextBugger. This can be attributed to the fact that masking in Data Augmentation allows the model to reconstruct overall contextual semantics rather than relying on specific words, directly countering word-substitution strategies. In contrast, masking in Random Smoothing conceals regions affected by character-level perturbations, enabling the model to focus on unperturbed input portions and mitigate such attacks.
Our method demonstrates effectiveness across all four attack types. During training, inserting [MASK] tokens at the beginning of clean sequences encourages the model to rely more on contextual semantics, enhancing its resilience to adversarial perturbations. During inference, replacing the lowest-frequency words with [MASK] removes potential attack points, such as rare substitutions or misspellings, while leveraging the model’s ability to reconstruct masked tokens using contextual information. This dual-phase masking strategy combines robust training with targeted inference adjustments, effectively defending against both word-level attacks (e.g., TextFooler, BERT-Attack) and character-level attacks (e.g., DeepWordBug, TextBugger).}

\section{Conclusion}\label{sec:conclusion}
{
In this study, we introduced a novel adversarial defense method that leverages the strategic insertion/replacement of [MASK] tokens both during classifier training and as a defensive action in response to attacks. This approach capitalizes on the vulnerability of less frequent tokens, which are often the adversarial manipulations. By training the classifier with samples that begin with [MASK] tokens and replacing the least frequent tokens in adversarial samples with [MASK], we strengthen the classifier's ability to detect and invalidate threats that traditional methods might miss. 
Our experimental results across various datasets and attack models have demonstrated that this method consistently outperforms existing defense approach, highlighting its effectiveness in enhancing the robustness of NLP applications against adversarial attacks. Additionally, the application of our method to Large Language Models has shown a significant improvement in their robustness, suggesting that this approach is not only effective but also adaptable to different NLP frameworks and applications. 
}

\begin{acknowledgments}
The authors would like to thank anonymous reviewers for their valuable suggestions to improve the quality of the article. This work is partially supported by the Australian Research Council Discovery Project (DP210101426), the Australian Research Council Linkage Project (LP200201035), AEGiS Advance Grant(888/008/268, University of Wollongong), and Telstra-UOW Hub for AIOT Solutions Seed Funding (2024, 2025).
\end{acknowledgments}

\starttwocolumn
\bibliographystyle{compling}
\bibliography{ref}

\end{document}